\documentclass{article}

\usepackage[accepted]{icml2026}

\usepackage[utf8]{inputenc} 
\usepackage[T1]{fontenc}    
\usepackage{hyperref}       
\usepackage{url}            
\usepackage{booktabs}       
\usepackage{amsfonts}       
\usepackage{nicefrac}       
\usepackage{microtype}      
\usepackage{lipsum}
\usepackage{fancyhdr}       
\usepackage{graphicx}       
\graphicspath{{media/}}     
\usepackage{xcolor}
\usepackage{amsmath}
\usepackage{subcaption}
\usepackage{comment}

\begin{document}

\twocolumn[
\icmltitle{Multimodal Hidden Markov Models for Persistent Emotional State Tracking}




\begin{icmlauthorlist}
\icmlauthor{Anamika Ragu}{comp}
\icmlauthor{Aneesh Jonelagadda}{comp}
\end{icmlauthorlist}

\icmlaffiliation{comp}{Kaliber AI, San Mateo, California, USA}

\icmlcorrespondingauthor{Aneesh Jonelagadda}{a.jonelagadda@kaliber.ai}

\icmlkeywords{Bayesian Nonparametric HMMs, Multimodal Question Answering, Factorial HDP-HMMs, Affective Computing, Conversational Emotion Recognition, Compute-Efficient Multimodal Fusion, LLM-as-a-Judge}
]



\printAffiliationsAndNotice{}  

\begin{abstract}
Tracking an interpretable emotional arc of a conversation via the sentiment of individual utterances processed as a whole is central to both understanding and guiding communication in applied, especially clinical, conversational contexts. Existing approaches to emotion recognition operate at the utterance level, obscuring the persistent phases that characterize real conversational dynamics. We propose a lightweight framework that models conversational emotion as a sequence of latent emotional regimes using sticky factorial HDP-HMMs over multimodal valence-arousal representations derived from simultaneous video, audio and textual input. We evaluate the quality of regime prediction using LLM-as-a-Judge, geometric, and temporal consistency metrics, demonstrating that the sticky HDP-HMM produces more interpretable regime sequences than the baseline Gaussian HMM at a fraction of the computational cost of LLM-based dialogue state tracking methods. In addition, Question-Answer experiments in a clinical dataset suggest that meaningful emotional phases can reliably be recovered from  multimodal valence-arousal trajectories and used to improve the quality of LLM responses in unstable affective regimes via context augmentation. This framework thus opens a path toward interpretable, lightweight, and actionable analysis of conversational emotion dynamics at scale.
\end{abstract}

\section{Introduction}

In conversational settings, being able to address a speaker’s emotional needs in a timely and nuanced manner is strongly linked to effective communication; in clinical contexts, this is further associated with improved care outcomes and therapeutic alliance. As such, research interest in dialogue state tracking and emotion recognition within conversational applications has grown substantially over the past decade, especially in the clinical context \cite{poria2019emotion}. Realistically, clinicians are responding more to evolving patterns of distress, reassurance, and engagement over the course of an interaction, rather than single utterances. Thus, when emotion is computationally inferred at the level of short utterances, producing sequences of labels that lack an explicit notion of persistence (as is the prevailing convention), the temporal-structural property of real conversations is obscured \cite{lee2022grayscale}. This motivates our first research question: to what extent can we detect stable and persistent regimes of emotion over the course of a conversation?

Moreover, emotion recognition systems have largely relied on discrete categorical labels (e.g. happy, sad, angry), reflecting annotative convenience rather than the underlying continuity of affect \cite{lee2022grayscale}. While valence-arousal (VA) representations offer a more psychologically grounded alternative, there has been comparatively little work on modeling the higher-level temporal dynamics of these representations using purely numerical methods. This raises our second research question: to what extent can continuous emotional trajectories be modeled and interpreted in a way that is both temporally coherent and useful for a conversational agent?

In this work, we address these questions by framing conversation in time-series as a sequence of latent emotional regimes. We propose a lightweight, multimodal framework that segments continuous VA trajectories into persistent states, leveraging Hidden Markov Models (HMMs), which are well-suited for modeling sequential data with underlying latent structural and temporal dependencies. While standard Gaussian HMMs provide a natural baseline for such classification tasks, they tend to over-segment emotional trajectories, producing unrealistically rapid switching between states. To address this, we employ a truncated sticky factorial HDP-HMM, which introduces an explicit bias toward self-transitions, allows the number of active regimes in a conversation to be inferred from data rather than fixed \textit{a priori}, and flexibly merges simultaneous multimodal inputs into coherent regimes.

\begin{figure}
    \centering
        \includegraphics[width=\linewidth, keepaspectratio]{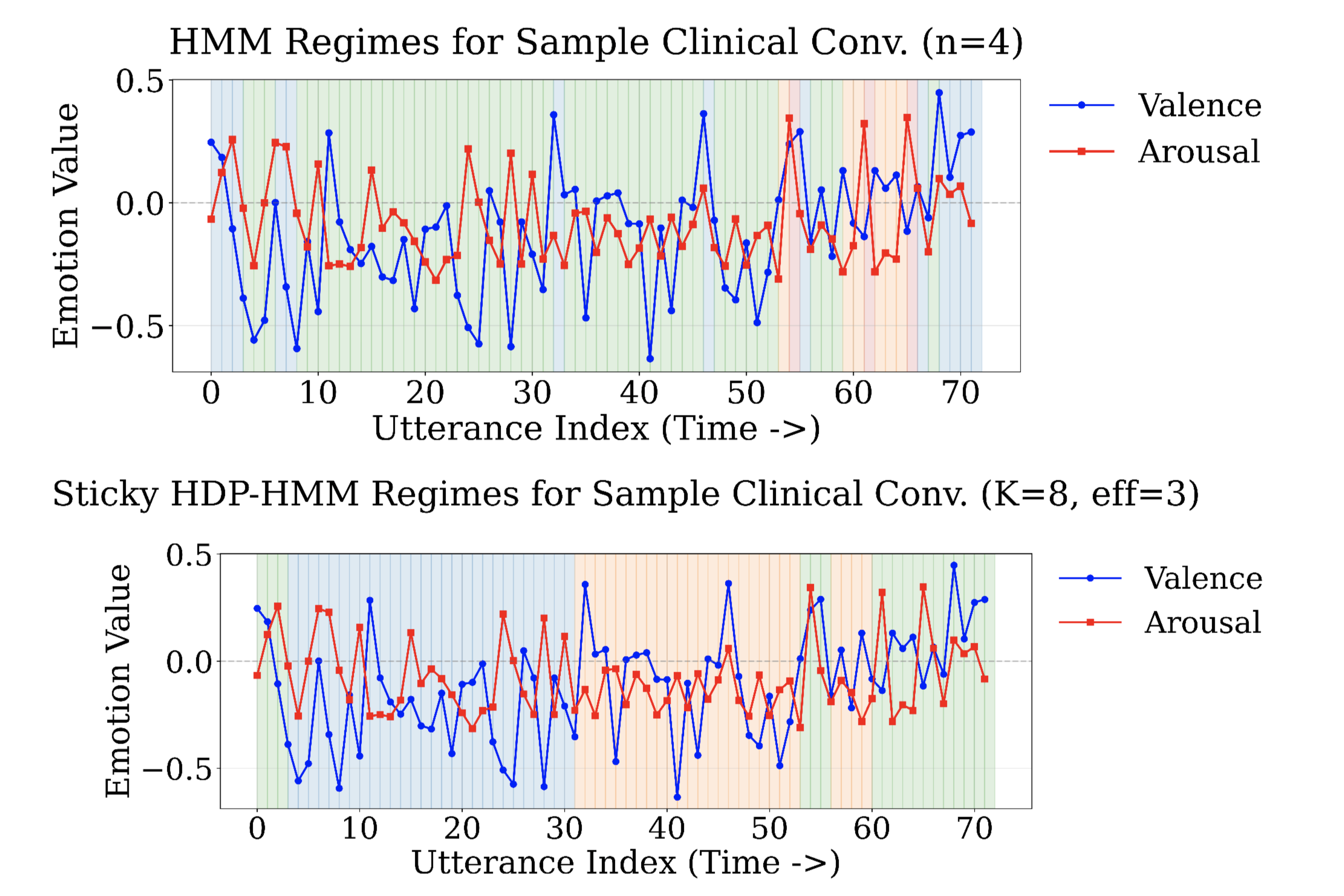}
    \caption{Comparison of conversational textual Valence-Arousal regimes identified by Gaussian HMMs (top) vs. Sticky HDP-HMMs (bottom). Regimes denoted by shading different colors across respective utterance indices. HMM specified to have 4 regimes (n=4), and Sticky-HDP-HMMs set to 8 maximum regimes (K=8), with 3 effective regimes identified.}
    \label{fig:sticky_gaussian_comparison}
\end{figure}

We show that continuous multimodal emotion trajectories can be segmented into stable, interpretable emotional regimes using sticky factorial HDP-HMMs (Figure \ref{fig:sticky_gaussian_comparison}), yielding coherent conversational structure without a reliance on large language model (LLM) inference at runtime. Crucially, our approach operates directly on numerical VA representations derived from audio, visual, and textual signals, enabling efficient inference without reliance on expensive LLM calls at runtime.

Beyond this computational efficiency, we emphasize interpretability in dialogue state tracking. By aligning inferred regimes with intuitive \& labeled interpretations of the dynamics in a given dataset, we move toward structured representations and guidance of conversational dynamics. We further explore mapping these regimes as higher-level communication strategies for use by an LLM to guide the conversation, providing a meaningful application leap from low-level affective signals to conversationally meaningful interaction patterns. Responses from an LLM in a clinical Question-Answer setting are compared with and without access to the interpretable regimes, and we find that augmenting the LLM context with the calculated regime with interpretable states improves response quality. Since our method for inferring these regimes is lightweight and "online", we thus contribute an efficient method leveraging multimodal signals for enhancing question-answer dynamics.

\section{Background \& Related Works}

\subsection{Emotion Recognition Models}

As the field of multimodal emotion recognition has expanded, speech and text became central modalities. A major reason is that emotion is often carried not only in facial cues but also in prosody, lexical choice, and discourse context; more recent multimodal systems fuse these signals to improve robustness, but the dominant pipeline still often performs utterance-level prediction over independently scored segments rather than modeling longer-range affective structure \cite{ramaswamy}. Further, as an improvement from discrete representation of emotions, modeling emotion as continuous coordinates \cite{ong2006resilience} is especially useful to our study because it preserves graded differences between nearby states and creates a structured numeric space that is well suited to downstream sequential models. For agentic conversational applications in particular, such nuance is critical: distinguishing between different types and intensities of negative affect can directly inform response strategies and interventions \cite{ong2006resilience}.

\subsection{Modern Limitations of Dialogue State Tracking}

More recently in Dialogue-State-Tracking (DST), schema-driven and LLM-based DST has become the dominant high-performance paradigm, especially in open-domain and zero-shot settings, because LLMs can generate slot values or state descriptions directly from dialogue history. However, this gain in flexibility comes at a substantial computational cost, and it does not fully solve the underlying robustness problem: multimodal and noisy dialogues still require a principled way to map heterogeneous inputs into stable state estimates \cite{carranza2025interpretable}. This is why DST is increasingly treated as a problem of robust state inference under uncertainty, especially when the dialogue's crucial signals have important temporal dimensions and are only partially observable in the presence of noisy data \cite{balaraman-etal-2021-recent}.

\subsection{Hidden Markov Models}
Hidden Markov Models (HMMs) provide a natural framework for modeling sequential data with latent structure, and their conceptual appeal lies in their establishment of unobserved states that govern the behavior of observed phenomena over time. The notion of inferring the internal states of time-series data from noisy observations allows us to treat conversational emotion as contiguous regimes with persistent affective dynamics and not as independent signals. Standard Gaussian HMMs, however, exhibit significant limitations when applied to granular sequential data such as utterance-level emotions. Maximum likelihood estimation in the model tends to favor rapid switching between states, leading to the over-segmentation of utterances, in which regimes fluctuate wildly at nearly every timestep. Attempts to stabilize these models by manually fixing the number of states or regularizing transitions often introduce rigidity, collapsing the probabilistic model into a more deterministic system and limiting its ability to adapt to the natural variability of conversational structure \cite{fox}. These limitations have motivated our development of more flexible models which can effectively adapt to real-world conversations.

\section{Methods}

\subsection{Utterance-level Emotion extraction}

Representing emotional state across two continuous dimensions capturing polarity (positive vs. negative) and activation (high vs. low energy) respectively \cite{russell1980circumplex}, the VA estimates produced at the utterance level serve as the foundational observations for subsequent temporal modeling and multimodal analysis of a given conversation. We consider text, audio, and video-based multimodal emotion detection from the textual sentiment, facial expressions, and aural tone of voice respectively. 

VA scores for written transcripts are produced by a fine-tuned DistilBERT model adapted for continuous regression \cite{NPC-Valence-Arousal-Prediction}. Specifically, the model maps each utterance to a two-dimensional Valence–Arousal space following Russell’s circumplex framework \cite{russell1980circumplex}. Transcripts fed to the model are segmented at the utterance level using time-aligned boundaries; additionally, we remove disfluencies and non-speech artifacts within the text. After this preprocessing step, each cleaned segment is passed independently to the model for inference.

VA scores for aural data are produced by a pruned Wav2Vec 2.0 model fine-tuned for dimensional speech emotion recognition \cite{wagner} on MSP-Podcast \cite{busso2025msp}, a large collection of natural conversational speech that is richly annotated with continuous VA-dominance scores. With raw 16 kHz audio as input, a regression head is applied to pooled transformer representations to generate these continuous outputs. Wav2Vec2.0 is particularly well-suited for this task because it captures prosodic and acoustic features—such as pitch, energy, and temporal dynamics—that carry affective information independently of lexical content. 

VA scores for visual data are produced by EmoNet, a facial affect model trained to estimate continuous valence and arousal from face images captured in naturalistic conditions \cite{toisoul2021emonet}. In our pipeline, video is first segmented at the utterance level and represented as a sequence of extracted frames.

\subsection{Multimodal feature construction}
\label{sec:multimodal_methods}

For each utterance $t$, let $\mathbf{x}^{\text{txt}}_t = (v^{\text{txt}}_t, a^{\text{txt}}_t) \in \mathbb{R}^2$ and $\mathbf{x}^{\text{aud}}_t = (v^{\text{aud}}_t, a^{\text{aud}}_t) \in \mathbb{R}^2$ denote the text and audio VA estimates, respectively. Rather than concatenating these features into a single observation vector, we treat each modality as a distinct observation channel generated by a shared latent emotional regime.

Formally, at each time step $t$, the observation is represented as a collection of modality-specific variables:
\begin{equation}
    \mathcal{X}_t = \left\{ \mathbf{x}^{\text{txt}}_t,\; \mathbf{x}^{\text{aud}}_t \right\}.
\end{equation}

Given a latent regime $z_t$, we assume that modalities are conditionally independent:
\begin{equation}
    p(\mathcal{X}_t \mid z_t) = p(\mathbf{x}^{\text{txt}}_t \mid z_t)\; p(\mathbf{x}^{\text{aud}}_t \mid z_t).
\end{equation}

Where each modality is modeled with its own Gaussian emission distribution.
%

This factorized emission model allows for each modality to exhibit distinct noise characteristics and variability within the same latent regime, avoiding the need to model a potentially complex joint distribution over linearly concatenated features. The overall log-likelihood decomposes additively across modalities:
\begin{equation}
    \log p(\mathcal{X}_t \mid z_t) = \log p(\mathbf{x}^{\text{txt}}_t \mid z_t) + \log p(\mathbf{x}^{\text{aud}}_t \mid z_t).
\end{equation}

This formulation also naturally extends directly to additional modalities. We incorporate visual affect by introducing a third modality $\mathbf{x}^{\text{vid}}_t = (v^{\text{vid}}_t, a^{\text{vid}}_t) \in \mathbb{R}^2$, computed from frame-level estimates and aggregated at the utterance level. The observation at time $t$ becomes
\begin{equation}
    \mathcal{X}_t = \left\{ \mathbf{x}^{\text{txt}}_t,\; \mathbf{x}^{\text{aud}}_t,\; \mathbf{x}^{\text{vid}}_t \right\},
\end{equation}
and the conditional independence assumption extends as
\begin{equation}
    p(\mathcal{X}_t \mid z_t) = \prod_{m \in \{\text{txt},\,\text{aud},\,\text{vid}\}} p(\mathbf{x}^{(m)}_t \mid z_t).
\end{equation}
All modality streams are standardised independently to zero mean and unit variance prior to modeling to ensure comparable scaling across modalities.

\subsection{Temporal regime modeling}

\subsubsection{Gaussian HMM baseline}
\label{sec:gaussian_hmm}

A standard Gaussian Hidden Markov Model \cite{hmm} provides a tractable baseline for latent regime detection. Given a conversation represented as a sequence of $T$ utterance-level observations $\mathbf{X} = \{\mathbf{x}_1, \dots, \mathbf{x}_T\}$, the model assumes that each observation is generated by a discrete latent state $z_t \in \{1, \dots, K\}$ evolving according to a first-order Markov process:
\begin{align}
    z_1 &\sim \boldsymbol{\pi}, \\
    z_t \mid z_{t-1} &\sim \mathrm{Categorical}(\mathbf{A}_{z_{t-1}}), \\
    \mathbf{x}_t \mid z_t = k &\sim \mathcal{N}(\boldsymbol{\mu}_k, \boldsymbol{\Sigma}_k),
\end{align}
where $\boldsymbol{\pi}$ is the initial state distribution, $\mathbf{A} \in [0,1]^{K \times K}$ is the transition matrix, and $(\boldsymbol{\mu}_k, \boldsymbol{\Sigma}_k)$ are the mean and covariance of the Gaussian emission for state $k$.

Model parameters are estimated via Expectation-Maximization, and the most probable latent sequence is recovered by Viterbi decoding. The number of states $K$ is fixed as a hyperparameter. We use a tied covariance structure ($\boldsymbol{\Sigma}_k = \boldsymbol{\Sigma}\ \forall k$) to reduce overfitting given the limited number of utterances per consultation.

\subsubsection{Sticky HDP-HMM}
\label{sec:hdphmm}

The standard Gaussian HMM requires $K$ to be specified in advance and imposes no prior preference for temporal persistence, leaving the model free to switch states at every step, as VA scores are independent across utterances with no temporal dimension. We address both limitations with a sticky Hierarchical Dirichlet Process HMM (sticky HDP-HMM) \cite{fox}, which infers the effective number of states from data and places an explicit self-transition bias that encourages temporally coherent regime occupancy.

\paragraph{HDP prior over transition distributions.}

The HDP-HMM places a hierarchical nonparametric prior over the rows of the transition matrix. A global mixing measure $\boldsymbol{\beta} \sim \mathrm{GEM}(\gamma)$ is drawn from a stick-breaking process with concentration $\gamma$, and each state-specific transition distribution is drawn as:
\begin{equation}
    \boldsymbol{\pi}_k \sim \mathrm{DP}\!\left(\alpha,\, \boldsymbol{\beta}\right), \quad k = 1, 2, \dots
\end{equation}
where $\alpha$ is a per-state concentration parameter. Because all rows share the same base measure $\boldsymbol{\beta}$, states that are never visited receive negligible probability mass, and the effective number of active states is inferred rather than presupposed.

\paragraph{Sticky self-transition bias.}

The standard HDP-HMM does not penalize rapid state switching. The sticky extension \cite{fox} augments each transition distribution with an additional mass $\kappa > 0$ on the self-transition:
\begin{equation}
    \boldsymbol{\pi}_k \sim \mathrm{DP}\!\left(\alpha + \kappa,\, \frac{\alpha \boldsymbol{\beta} + \kappa \boldsymbol{e}_k}{\alpha + \kappa}\right),
\end{equation}
where $\boldsymbol{e}_k$ is the unit vector on state $k$. Increasing $\kappa$ raises the prior probability of remaining in the current state, directly encoding the assumption that emotional regimes are persistent rather than fleeting. This is particularly appropriate for clinical conversation, where affective phases typically span multiple consecutive utterances.

\paragraph{Truncated inference.}

Exact inference under the HDP-HMM is intractable. We employ a truncated weak-limit approximation \cite{blei,Ishwaran01032001}, fixing an upper bound $K_{\max}$ on the state space while allowing the posterior to concentrate on a smaller active subset. $K_{\max}$ is informed by the number of states at which the quality of Gaussian HMM modeling degrades, as determined by hyperparameter tuning (see Appendix-\ref{hyperparameter-tuning}).

Inference proceeds via collapsed Gibbs sampling, alternating between sampling the latent state sequence $z_{1:T}$ via the forward-backward algorithm and resampling the model hyperparameters $(\alpha, \kappa, \gamma)$ and emission parameters $\{(\boldsymbol{\mu}_k, \boldsymbol{\Sigma}_k)\}_{k=1}^{K_{\max}}$ from their conjugate posteriors. The most probable state sequence for reporting is taken as the Viterbi path under the posterior mean parameters.

\paragraph{Emission model.}

As with the Gaussian HMM baseline, each active state $k$ is associated with a Gaussian emission $\mathcal{N}(\boldsymbol{\mu}_k, \boldsymbol{\Sigma}_k)$ over the $d$-dimensional observation space. Because the emission structure is otherwise identical to the baseline, any differences in regime quality between the two models are attributable solely to the nonparametric prior and the sticky self-transition mechanism.

\subsection{Evaluation}
\label{sec:evaluation}

Evaluating unsupervised temporal segmentation quality is non-trivial in the absence of ground-truth regime annotations. We employ an LLM-as-a-Judge to evaluate our methodology by constructing a reference label using a large language model (LLM) prompted to segment each conversation into coherent emotional phases and assign a descriptive label to each. Formally, given a conversation comprising $T$ utterances $\mathbf{u} = (u_1, u_2, \ldots, u_T)$, the LLM assigns each utterance a regime label, producing a reference label sequence $\mathbf{r} \in {1, \ldots, K}^T$ over $K$ distinct labels, where $K$ is not fixed in advance. The prompt provides the full utterance-level transcript and instructs the model to identify utterances sharing a consistent affective character; labels may recur freely across the sequence, permitting regimes to alternate and re-emerge. The prompt instructs the model to identify transitions significant enough to require a clinician to meaningfully change their therapeutic approach, with regimes consolidated to a maximum of 8 distinct labels per consultation. Reference labels were generated using GPT-5.4 via the OpenAI API, and the full prompt is available in Appendix-\ref{comparison-llm-prompt}.

These LLM-derived labels are treated as approximate ground truth for the purposes of quantitative evaluation; our sticky HDP-HMM independently produces a predicted label sequence $\mathbf{p} \in {1, \ldots, M}^T$, where the number of predicted regimes $M$ need not equal $K$. Because both labels are produced independently, their cluster indices are arbitrary — predicted label $i$ carries no semantic correspondence to reference label $j$. This is the label permutation problem. Predicted regime sequences are therefore aligned to reference labels via the Hungarian algorithm \cite{hungarian} to resolve this, with details in Appendix-\ref{hungarian-algo}. 

We report segment-level F1, boundary F1 (within a tolerance of $\pm 1$ utterance), and normalized mutual information (NMI) between predicted and reference label sequences. In addition, we evaluate regime stability directly via the mean regime duration (in utterances) and the proportion of single-utterance regimes, both of which should be low for a temporally coherent model.

For Question-Answer augmentation evaluation, we construct a paired evaluation over mid-conversation prompts drawn from the PriMock57 dataset \cite{papadopoulos-korfiatis-etal-2022-primock57}. For each consultation, the first half of the dialogue is provided as context, and the LLM is tasked with generating the next response in the interaction.

We compare two conditions: (1) \textbf{Baseline}: The LLM is prompted with the conversation history only. (2) \textbf{Regime-augmented}: The LLM is prompted with the same conversation history, along with a compact structured summary of the current inferred emotional regime (valence, arousal, regime label, persistence, and consultation phase), derived from the multimodal sticky HDP-HMM.

No additional task-specific instructions are introduced beyond this structured
signal. The generation model is GPT-5.4, with three candidate responses
generated per condition and a canonical response selected via a separate model call.
Evaluation uses an LLM-as-a-judge framework \cite{zheng2023judging} also implemented
with GPT-5.4, comprising two complementary judgment types: pairwise A/B preference
(with randomized condition-to-label assignment to control for position bias) and
absolute rubric scoring across four dimensions. Ground truth is defined as the next
three actual clinician turns in the transcript, used as a reference anchor for the
judge. For each prompt, the judge model is asked to select the better response between baseline and regime-augmented across multiple criteria: Contextual Appropriateness, Emotional Attunement, Helpfulness, and Conversational Coherence. Full prompts for all LLM calls are provided in Appendix~\ref{qnallm-prompt}. 

\section{Results}

The results are organized around five claims: that HMMs can segment clinical conversations into coherent emotional regimes (Section \ref{sec:results_segmentation}); that the sticky HDP-HMM produces meaningfully better regime structure than the Gaussian HMM baseline (Section \ref{sec:results_sticky}); that the framework extends to multimodal observation spaces (Section \ref{sec:results_multimodal}); that these recovered regimes are interpretable in affective and clinical terms (Section \ref{sec:results_interpretability}); and that downstream augmentation of a QnA task LLM with the current interpreted regime significantly improves the quality of generated conversational responses (Section \ref{sec:results_qa}).

\subsection{HMM Regime Segmentation Baseline}
\label{sec:results_segmentation}

As a prerequisite for all subsequent claims, latent state models must recover structured regime sequences from utterance-level VA trajectories. The key question is whether the resulting segmentation reflects genuine conversational structure rather than noise.

Across all 57 PriMock57 consultation transcripts alone, both the Gaussian HMM and the sticky HDP-HMM recover regime sequences with sustained occupancy. The Gaussian HMM produces a mean regime duration of 3.04 utterances, while the sticky HDP-HMM produces that of 9.66 utterances. Both are substantially above the duration of 1.0 expected under random assignment. Representative decoded sequences show that latent states persist across contiguous conversational spans. The emission means recovered by both models occupy distinct and interpretable regions of the VA plane. Across consultations, states partition the VA space into recognizable affective quadrants, indicating that the latent states capture meaningful affective modes. The positive mean inter-regime centroid (Euclidean) distance of 0.469 with the Gaussian HMM and 0.339 with the sticky HDP-HMM further supports this separation, indicating that both models identify geometrically distinct affective clusters.

\subsection{Sticky HDP-HMM Comparison to Gaussian HMM}
\label{sec:results_sticky}

Both models recover non-trivial regime structure, but differ in how realistically they represent temporal dynamics. The sticky HDP-HMM captures temporally coherent phases that reflect persistent affective structure, whereas the Gaussian HMM over-segments what may be real emotional states in conversation, generating more noise than interpretable insights. Table \ref{tab:temporal_coherence} summarizes temporal coherence metrics across all consultations. The sticky HDP-HMM reduces single-utterance regimes by nearly one full order of magnitude and reduces regime shifts by a factor of four, while achieving substantially longer regime durations. These differences indicate a strong improvement in temporal coherence.

\begin{table}[h]
\centering
\caption{Quantitative metrics averaged across 57 PriMock57 consultations. Better values are \textbf{bolded}.}
\label{tab:temporal_coherence}
\resizebox{\columnwidth}{!}{

\begin{tabular}{lcc}
\toprule
\textbf{Metric} & \textbf{Gaussian HMM} & \textbf{Sticky HDP-HMM} \\
\midrule
Mean regime duration (utterances) $\uparrow$ & 3.04 & \textbf{9.66} \\
Single-utterance regime fraction $\downarrow$ & 0.561 & \textbf{0.061} \\
Regime shifts per consultation $\downarrow$ & 21.11 & \textbf{5.33} \\
Mean temporal purity $\uparrow$ & 0.384 & \textbf{0.850} \\
Intra-regime VA variance $\uparrow$ & 0.0318 & \textbf{0.0619} \\
Inter-regime centroid distance & \textbf{0.469} & 0.339 \\
Mean log-likelihood  $\uparrow$ & \textbf{45.61} & 44.19 \\
LLM-as-a-Judge: Mean Segment-Level F1 $\uparrow$ & 0.0434 & \textbf{0.0770} \\
LLM-as-a-Judge: Mean Boundary F1 $\uparrow$ & 0.266 & \textbf{0.302} \\
LLM-as-a-Judge: Mean NMI $\uparrow$ & 0.229 & \textbf{0.431} \\
\bottomrule
\end{tabular}}
\end{table}

Notably, the Gaussian HMM achieves marginally higher mean log-likelihood (45.61 vs.\ 44.19), suggesting a higher accuracy upon first glance. This is expected due to the objective of log-likelihood optimizing the observation space without temporal regularization. The difference is small relative to the large gains in temporal coherence, and log-likelihood alone is not a reliable indicator of regime quality in this setting. For this reason, we extend our evaluation methodology to include LLM-as-a-Judge, and find that the sticky HDP-HMM substantially outperforms the Gaussian HMM across all three reference-based metrics. The LLM judge assigned a mean of 3.61 distinct regimes per consultation with an average duration of 14.18 utterances, a granularity considerably closer to the sticky model's behavior than the Gaussian's.

The most compelling result is Normalized Mutual Information (NMI), where the sticky model scores 0.431 against 0.229, winning on 48 of 57 consultations — indicating that its regime assignments are substantially more aligned with the LLM judge's partition of the affective space. Boundary F1 corroborates this, with the sticky model scoring 0.302 against 0.266. Segment-level F1 is low for both models, which is expected given the metric's requirement for exact coincidence of both boundaries and labels; we therefore treat NMI and boundary F1 as the primary judge-based metrics.

Taken together, these results confirm that the Gaussian HMM's lack of a temporal persistence prior causes it to over-segment. The sticky HDP-HMM, in contrast, recovers sustained affective structure, and we conclude that temporal regularization via the sticky prior is essential for detecting clinically meaningful regime shifts in this setting.

\subsection{Extension to multimodal observation spaces}
\label{sec:results_multimodal}

We first establish the collinearity of the two streams before evaluating the combined model to assert our assumption of conditional independence of modalities' emission channels (Section-\ref{sec:multimodal_methods}), and find the Pearson correlation between text valence and audio valence to be $r = 0.361$, and between text arousal and audio arousal to be $r = 0.188$. These values indicate a weak relationship; this divergence reflects the known phenomenon of semantic-prosodic decoupling in clinical speech, where patients frequently describe distressing content in a controlled vocal register \cite{cummins, schuller2013computational}. Table~\ref{tab:modality_comparison} summarizes regime statistics across all three configurations of the sticky HDP-HMM. 

\begin{table}[h]
\centering
\caption{Sticky HDP-HMM regime statistics by modality configuration. Best values are \textbf{bolded}.}
\label{tab:modality_comparison}
\resizebox{\columnwidth}{!}{
\begin{tabular}{lccc}
\toprule
\textbf{Metric} & \textbf{Text (2D)} & \textbf{Audio (2D)} & \textbf{Combined (4D)} \\
\midrule
Effective regimes $\uparrow$ & \textbf{4.46} & 3.25 & 3.53 \\
Regime shifts $\downarrow$ & 5.33 & \textbf{4.00} & 8.05 \\
Dominant regime share & 0.48 & 0.57 & 0.49 \\
Transition entropy $\downarrow$ & 0.61 & 0.59 & \textbf{0.49} \\
Mean regime duration $\uparrow$ & 9.66 & \textbf{12.82} & 7.50 \\
Single-utterance fraction $\downarrow$ & 0.06 & \textbf{0.03} & 0.09 \\
\bottomrule
\end{tabular}}
\end{table}

The results reveal a clear modality-specific trade-off. The text-only model recovers the richest affective structure, with the highest effective regime count (4.46) and a comparatively balanced dominant regime share (0.48), suggesting it captures a wider repertoire of emotional states. The audio-only model, by contrast, favors temporal stability: it produces the fewest regime shifts (4.00), the longest mean regime duration (12.82 utterances), and the lowest single-utterance fraction (0.03), consistent with the smoothly-varying nature of prosodic signals in clinical conversation. The combined 4D model presents a mixed picture. It achieves the lowest transition entropy (0.49), indicating that its between-regime dynamics are the most structured and predictable of the three configurations, yet it also exhibits the highest regime-shift rate (8.05) and shortest mean duration (7.50), and the largest single-utterance fraction (0.09). This apparent tension is interpretable: when text and audio disagree — as the low inter-modal correlations suggest they frequently do — the model must resolve conflicting emission signals, which can induce brief excursions into transient regimes that would not arise under either unimodal model alone. 

As we know, the model scales transparently to an arbitrary number of modality channels, where adding a new modality requires only the specification of an additional emission distribution $p(\mathbf{x}_t^{(m)} \mid z_t)$ for that channel; the HDP prior over the transition matrix and the inference procedure remain unchanged. We exploit this flexibility to extend the evaluation to a trimodal setting using MELD \cite{poria2019meld}, a multimodal corpus of emotionally labeled dialogue segments drawn from the television series \textit{Friends}, where text, audio, and video valence-arousal streams are all available.

Table~\ref{tab:6D_modality_comparison} summarizes regime statistics across all four configurations on MELD. The unimodal models are broadly comparable in effective regime count (2.00–2.30) and show near-zero single-utterance fractions, indicating stable, persistent affective states within each modality. The combined 6D model highlights the framework’s flexibility to integrate richer multimodal signals, yielding a more expressive affective decomposition (2.70 effective regimes, dominant regime share of 0.60). This added granularity comes with brief transitions that capture transient disagreements between modalities. Consistent with the 4D setting, this behavior demonstrates adaptive segmentation, where increased input complexity leads to finer-grained, but not noisy, affective structure.

\begin{table}[h]
\centering
\caption{Sticky (Factorial) HDP-HMM regime statistics by modality configuration on MELD.}
\label{tab:6D_modality_comparison}
\resizebox{\columnwidth}{!}{
\begin{tabular}{lcccc}
\toprule
\textbf{Metric} & \textbf{Text (2D)} & \textbf{Audio (2D)} & \textbf{Video (2D)} & \textbf{Combined (6D)} \\
\midrule
Effective regimes $\uparrow$           & 2.00 & 2.10 & 2.30 & 2.70 \\
Regime shifts $\downarrow$             & 0.90 & 1.10 & 1.10 & 2.70 \\
Dominant regime share                  & 0.75 & 0.66 & 0.74 & 0.60 \\
Transition entropy $\downarrow$        & 0.40 & 0.43 & 0.47 & 0.73 \\
Mean regime duration $\uparrow$        & 5.07 & 4.49 & 4.52 & 2.80 \\
Single-utterance fraction $\downarrow$ & 0.05 & 0.00 & 0.00 & 0.33 \\
\bottomrule
\end{tabular}
}
\end{table}

\subsection{Interpretation of Regimes}
\label{sec:results_interpretability}

Statistical coherence alone is insufficient; the recovered regimes must also correspond to meaningful affective states. We therefore analyze the geometry of emission means, their consistency across modalities, and their temporal realization in case studies.

\begin{figure*}[t]
    \centering
    \includegraphics[width=0.9\textwidth]{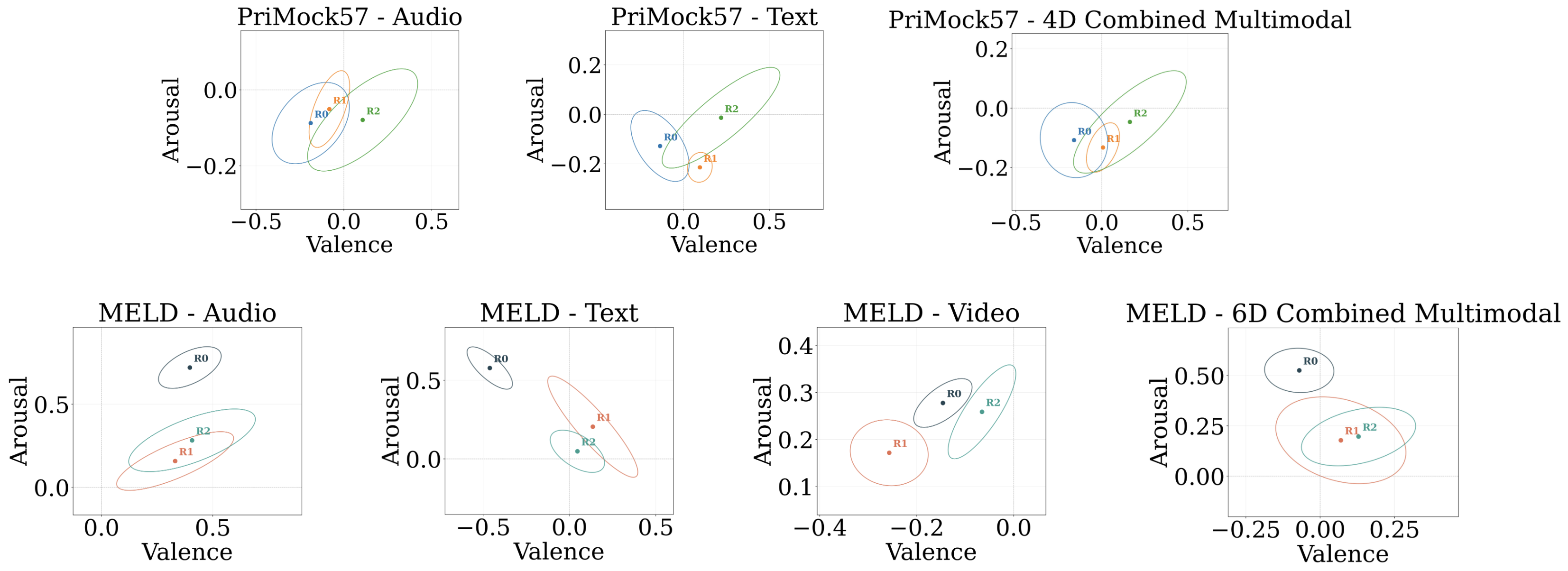}
    
    \vspace{0.5em}
    \caption{
    \textit{Multimodal regime emission structure (Gaussian) across datasets.} Each point denotes the mean VA centroid of an inferred regime, and each ellipse shows the corresponding 1-$\sigma$ emission covariance, representing the spread of utterance-level observations assigned to that regime. Regime labels (R0, R1, R2) are assigned in ascending order of valence. PriMock57 regimes cluster near neutral valence with low arousal, consistent with the affective range of primary care consultations. MELD regimes show greater separation and higher arousal, reflecting the more expressive emotional range of scripted TV dialogue.
    }
    \label{fig:regime_emissions}
\end{figure*}

Across both PriMock-57 and MELD, the learned regimes organize into a structured trajectory in VA space rather than forming arbitrary clusters. The elliptical structure visible in the text and audio emission plots as seen in Figure \ref{fig:regime_emissions}  supports this interpretation: individual modalities exhibit elongated, overlapping distributions, while the combined emissions collapse these into intermediate, interpretable distributions. The combined VA means lie between modalities, indicating that the factorial emission model resolves these discrepancies into a consensus representation rather than privileging a single modality. This behavior is critical for interpretability: regimes reflect shared affective structure while retaining modality-specific nuance.

Temporal structure further reinforces this interpretation. In PriMock-57 (day2\_cons10), the combined sticky HDP-HMM is dominated by the lowest-valence regime, with $R0$ occupying 61.3\% of utterances, compared with 21.0\% for $R1$ and 17.7\% for $R2$. The sequence begins with a brief high-valence segment ($R2$), then settles into a long dwell in the dominant negative regime ($R0$), before shifting to an intermediate-valence regime ($R1$) and ending with a short return to the highest-valence regime ($R2$). In MELD (dialogue 0), the early interaction is likewise governed by a single high-occupancy regime, but here that regime is the highest-valence state: $R2$ occupies 71.4\% of the clip, whereas $R1$ accounts for 21.4\% and $R0$ only 7.1\%. After this extended $R2$ segment, the clip transitions briefly into a low-valence, high-arousal state ($R0$) and then ends in a short intermediate-valence state ($R1$). These transitions occur in contiguous blocks of utterances, consistent with temporally extended affective phases. Importantly, regime-specific variance provides additional evidence that these states are meaningful. In PriMock-57, lower-valence regimes (R0, R1) exhibit tighter variance, while more positive regimes (R2) show increased dispersion, suggesting greater heterogeneity in affective expression. In MELD, high-arousal regimes are both extreme in mean VA and relatively stable within modalities, indicating distinct emotional peaks rather than noise-driven outliers.

Taken together, these findings show that the sticky factorial HDP-HMM recovers regimes that are geometrically structured, temporally persistent, and consistent across modalities. The factorized emission model is central to this interpretability: by explicitly modeling modality-specific contributions, it exposes systematic cross-modal differences while producing unified regime representations that align with coherent phases of interaction.

\subsection{Multimodal QnA Augmentation with Regimes}
\label{sec:results_qa}

We evaluate whether inferred emotional regimes can serve as a compact conditioning signal for large language models (LLMs) in clinical question-answer (QnA) settings. Figure~\ref{fig:qa_examples} shows a representative comparison. Regime-augmented responses tend to incorporate brief affective acknowledgments calibrated to the inferred emotional state before continuing the clinical thread, whereas baseline responses proceed directly to clinical content without tonal modulation.

\begin{figure}[H]
    \centering
    \fbox{
    \parbox{0.9\linewidth}{
    \textit{Inferred regime: negative valence with slightly elevated arousal
(prolonged fatigue/frustration), stable for 5 consecutive turns.} \\[6pt]
    \textbf{Baseline:} ``Yes, that can happen with this sort of illness. Have you noticed feeling mentally slower than usual? And how’s your sleep been?'' \\[3pt]
    \textbf{Regime-augmented:} ``That sounds really uncomfortable, especially with it keeping you up at night. Can you tell me whether the itch is worse at any particular time of day?'' 
    }}
    \caption{Representative comparison of baseline and regime-augmented responses. The augmented response adds a brief affective acknowledgment and manageable clinical inquiry.}
    \label{fig:qa_examples}
\end{figure}

\begin{table}[h]
\centering
\caption{Absolute rubric scores (mean, scale 1---5) under baseline and regime-augmented conditions across $N=57$ PriMock57 consultations. $p$-values from two-sided paired $t$-tests. Stratified results show regime-dependent effects. Better values are \textbf{bolded}.}
\label{tab:qa_eval}
\resizebox{\columnwidth}{!}{
\begin{tabular}{lcccc}
\toprule
\multicolumn{5}{c}{\textbf{Overall}} \\
\midrule
\textbf{Dimension} & \textbf{Baseline} & \textbf{Augmented} & \textbf{$\Delta$} & \textbf{$p$} \\
\midrule
Affective Attunement      & 1.87 & \textbf{2.07} & $+0.20$ & 0.104 \\
Clinical Appropriateness  & \textbf{4.04} & 4.02 & $-0.02$ & 0.866 \\
Contextual Coherence      & 3.95 & \textbf{4.10} & $+0.15$ & 0.106 \\
Specificity               & \textbf{4.00} & 3.88 & $-0.12$ & 0.082 \\
Ground Truth Alignment    & 2.09 & \textbf{2.28} & $+0.19$ & 0.076 \\
\midrule
\multicolumn{5}{c}{\textbf{Stratified by Regime Persistence}} \\
\midrule
\textbf{Condition} & \textbf{n} & \textbf{Win Rate} & \textbf{$p$} & \\
\midrule
\textbf{Unstable regimes ($\leq 5$ turns)} & 28 & \textbf{0.714} & \textbf{0.036} & \\
Stable regimes ($> 5$ turns)      & 11 & 0.636 & 0.549 & \\
\bottomrule
\end{tabular}}
\end{table}

Table~\ref{tab:qa_eval} summarizes absolute rubric scores across all $N=57$ consultations. Each dimension captures a distinct aspect of response quality: \textit{Affective Attunement} measures sensitivity to the patient's emotional state; \textit{Clinical Appropriateness} reflects adherence to plausibly sound medical practice; \textit{Contextual Coherence} assesses logical consistency with the preceding dialogue; \textit{Specificity} captures the precision and detail of the response; and \textit{Ground Truth Alignment} measures agreement with the reference clinician response. Across dimensions, the regime-augmented condition exhibits consistent directional improvements in affective attunement, contextual coherence , and ground truth alignment, though none reach conventional significance thresholds at this sample size. Clinical appropriateness remains unchanged, and specificity shows a small negative shift. Overall, these results suggest modest but consistent gains in affective and contextual alignment without degradation of core clinical quality.

Despite the lack of significance at the aggregate conversation level, stratified analysis by regime persistence and stability localizes the effect. Under rubric-based evaluation, augmentation yields a significant improvement in emotionally unstable consultations, but not in stable ones. This pattern is consistent with the hypothesis that regime summaries are most informative when the patient’s emotional state is shifting and not yet fully inferable from local context. These results provide qualified support for emotional regimes as a useful intermediate representation for conditioning LLM behavior in specific segments of conversational dialogues. In addition, unstable regimes are often when the criticality of giving an appropriate responses is high; steady-state conversation segments are lower complexity and often handled well by both the baseline and augmented models.

\section{Conclusion}

In this work, we presented a lightweight framework for modeling conversational emotion as a sequence of persistent latent regimes. By applying truncated sticky factorial HDP-HMMs to multimodal valence–arousal representations derived from audio and textual signals, we showed that it is possible to recover temporally coherent and interpretable affective structure that is obscured by standard utterance-level approaches. Relative to Gaussian HMM baselines, the proposed model produces substantially more stable regime sequences while retaining meaningful structure in the underlying affective space.

Beyond modeling, we explored the use of inferred regimes as a compact representation of conversational states for downstream systems. Using a clinical question-answer setting, we find that while incorporating regime-level summaries into LLM context has no significant impact on response quality for steady-state affective conversational trajectories, it \textit{does} significantly lead to better aligned responses in unstable affective regimes where the user is emotionally fluctuating.

This suggests a broader view of regime inference as a form of state compression, where it serves as a low-dimensional, persistent signal that allows LLMs to respond in alignment with conversational dynamics without repeatedly processing the full dialogue history. Such representations offer a practical pathway toward more efficient and stable multimodal question-answer systems.

Future work will focus on strengthening both evaluation and interpretability. On the evaluation side, replacing the LLM-as-a-Judge with human raters and developing datasets with regime annotations grounded in real cognition would enable more rigorous benchmarking of the proposed framework. On the interpretability side, projecting continuous VA trajectories as defined by regimes' emission distributions onto discrete emotion representations may yield regime sequences that are more directly legible to both clinicians and downstream sequences. Taken together, these directions build towards a broader vision of conversational emotion analysis that is simultaneously lightweight, interpretable across fields, and clinically actionable. 

\bibliography{references}
\bibliographystyle{icml2026}

\appendix
\section{Hungarian Algorithm for Label Alignment}
\label{hungarian-algo}
To match the LLM-as-Judge regime indices with our model prediction indices, we first construct an $M \times K$ cost matrix $\mathbf{C}$ whose entries measure the utterance-level disagreement between each pair of predicted and reference labels:

\begin{equation} 
    C_{ij} = -\sum_{t=1}^{T} [p_t = i]\cdot[r_t = j], \quad i \in {1,\ldots,M},\ j \in {1,\ldots,K} 
\end{equation}

so that $-C_{ij}$ counts the number of utterances on which predicted cluster $i$ and reference label $j$ co-occur. Setting $n = \max(M, K)$, we zero-pad $\mathbf{C}$ to an $n \times n$ matrix and find the optimal permutation $\pi^* \in S_n$ minimizing total assignment cost:

\begin{equation} 
    \pi^* = \underset{\pi \in S_n}{\arg\min} \sum_{i=1}^{n} C_{i,,\pi(i)} 
\end{equation}

where assignments involving padded dummy rows or columns incur zero cost and are discarded after solving. When $M \neq K$, $\pi^*$ therefore constitutes a partial assignment, matching each label in the smaller set to its most overlapping counterpart in the larger. The optimal $\pi^*$ equivalently maximizes total utterance-level overlap between predicted and reference labels. Applying $\pi^*$ to $\mathbf{p}$ yields a remapped prediction sequence $\tilde{\mathbf{p}}$ defined by $\tilde{p}_t = \pi^{(p_t)}$, which now shares a common label space with $\mathbf{r}$.

\section{Selection of $K_{\max}$ from Hyperparameter Tuning Gaussian HMM}
\label{hyperparameter-tuning}

\begin{figure}[t]
    \centering
    \includegraphics[width=0.9\linewidth]{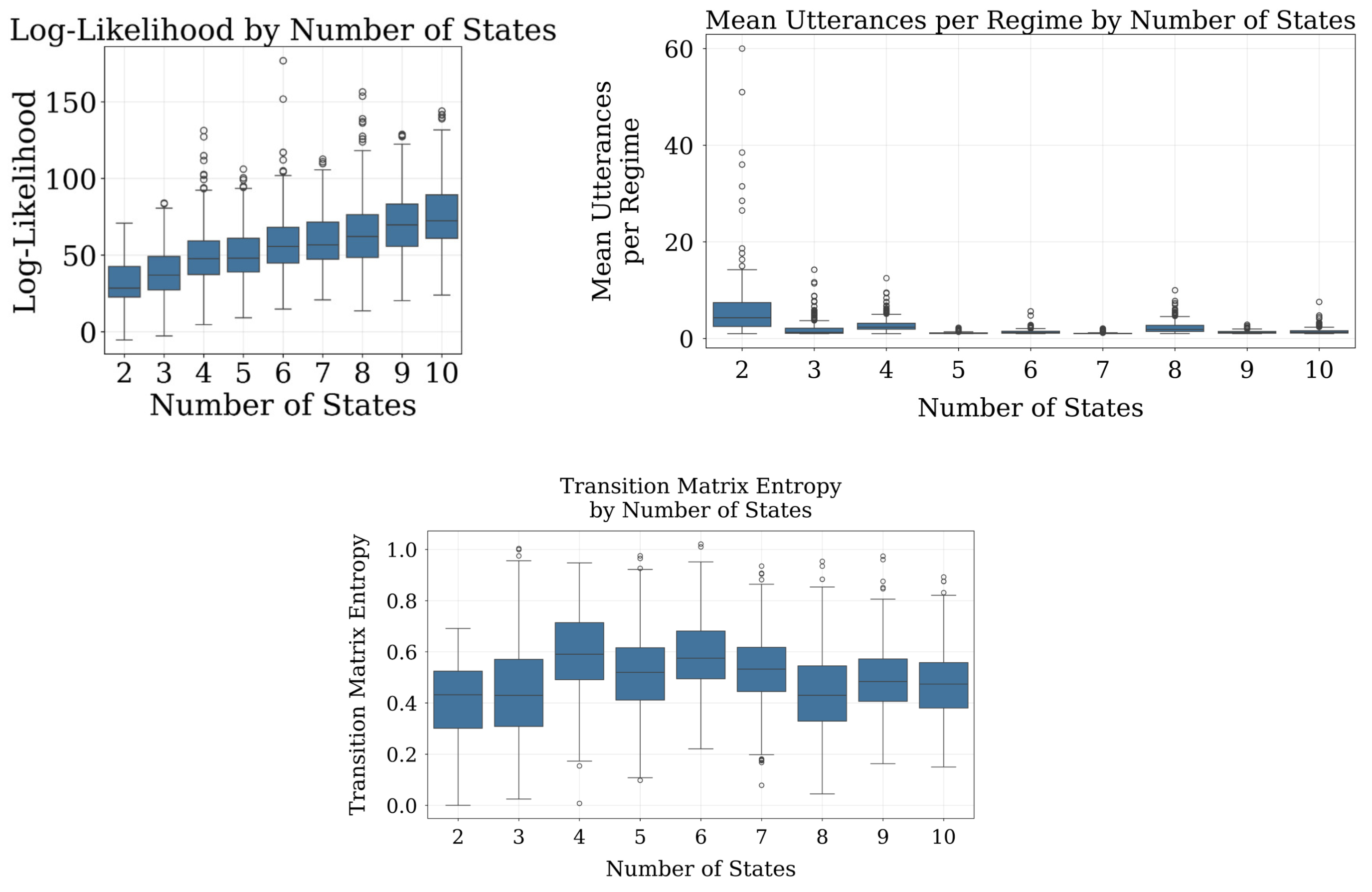}
    \caption{\textit{Model selection behavior across different numbers of hidden states.} Log-likelihood increases monotonically with $K$, but structural metrics such as mean regime duration and transition entropy indicate reduced interpretability beyond $K \approx 6$. Beyond $K=8$, regimes become increasingly short-lived and fragmented, motivating the choice of $K_{\max}=8$ as a balance between fit and temporal coherence.}
    \label{fig:k_selection}
\end{figure}

Figure~\ref{fig:k_selection} shows model behavior as the number of states increases, evaluated via log-likelihood, mean utterances per regime, and transition matrix entropy. While log-likelihood improves monotonically with additional states, this gain is accompanied by diminishing structural coherence: mean regime duration drops sharply beyond $K \approx 4$, and transition entropy remains high, indicating increasingly diffuse and less interpretable dynamics.

Notably, beyond $K=8$, additional states primarily capture short-lived or singleton regimes rather than stable affective structure, reflecting over-fragmentation of the observation space. Additionally, transition entropy rebounds after that point to increase. This aligns with the point at which improvements in Gaussian HMM fit are driven more by hyperparameter flexibility than by meaningful structure in the data. Accordingly, we set $K_{\max}=8$ as a balance between representational richness and temporal coherence.

\section{LLM Prompts}
\subsection{For comparison between Gaussian \& sticky HDP-HMMs:}
\label{comparison-llm-prompt}

This prompt is used to obtain LLM-based regime annotations for comparison with Gaussian and sticky HDP-HMM segmentations. All calls use GPT-5.4 with temperature $= 0.0$ to encourage deterministic, consistent labeling.

\begin{quote}\ttfamily\small
You are an expert clinical supervisor reviewing a patient's utterances from a therapy session.

You will be given a numbered list of the patient's utterances.

Your task is to segment the conversation into emotional regimes. A regime is a sustained affective state that persists across multiple utterances. A new regime only begins when the patient's emotional state shifts so substantially that a clinician would need to meaningfully change their approach to meet the patient's needs — for example, shifting from active listening to crisis intervention, or from psychoeducation to emotional validation.

Minor variations in wording, topic, or intensity within the same underlying emotional state do NOT constitute a new regime. A patient venting about different topics while remaining equally distressed is in one regime. A patient who moves from distress into flat detachment has crossed into a new one.

Rules:
- Use a STRICT MAXIMUM of 8 distinct regime labels total.\\
- Label each regime with a short phrase (2--5 words) describing the dominant affective character, e.g. "acute distress", "guarded withdrawal", "calm reflection", "resigned hopelessness".\\
- The same label may recur if the patient returns to a prior state.\\
- Every utterance must receive a label.

First, write 2--3 sentences describing the overall emotional arc of the conversation and identifying the major regime transitions you observe. Then output a JSON array, one object per utterance, in this exact format:\\
\texttt{[{"t": <utterance index>, "label": "<regime label>"}, ...]}

Output only the reasoning followed by the JSON. No other commentary.

Transcript:\\
\{transcript\}
\end{quote}

\subsection{QnA Task/Evaluation Prompts:}
\label{qnallm-prompt}

This section documents all prompts used in the QnA evaluation pipeline (Section~\ref{sec:results_qa}). All generation and evaluation calls use GPT-5.4. Judge calls use temperature $= 0.0$; generation calls use temperature $= 0.7$.

\subsubsection*{A.2.1 Response Generation — Baseline System Prompt}

\begin{quote}\ttfamily\small
You are an experienced primary care clinician conducting a medical consultation. Given the conversation history, generate the next clinician response. Be concise, empathetic, and clinically appropriate. Respond with one turn only.
\end{quote}

\subsubsection*{A.2.2 Response Generation — Regime-Augmented System Prompt}

The augmented system prompt appends the following block to the baseline prompt above, where \texttt{\{regime\_block\}} is instantiated per consultation:

\begin{quote}\ttfamily\small
During history-taking phases, use the following regime signal to calibrate empathic acknowledgment only. Reserve diagnostic and management responses for assessment phases.:
\{regime\_block\}
\end{quote}

The regime block itself is structured as follows, with fields derived from the multimodal sticky HDP-HMM at the consultation midpoint:

\begin{quote}\ttfamily\small
[Emotional Regime Summary]\\
Consultation phase: \{history-taking | assessment/management\}\\
Current regime: \{label\} (valence: \{v\}, arousal: \{a\})\\
Regime persistence: \{n\} consecutive turns (\{stable | unstable\})\\
Regime shifts so far: \{k\}
\end{quote}

Consultation phase is assigned as \textit{history-taking} when the midpoint turn index falls below 60\% of the total turn count, and \textit{assessment/management} otherwise. Stability is defined as persistence $> 5$ consecutive turns.

\subsubsection*{A.2.3 Response Generation — User Prompt (Both Conditions)}

The user prompt is identical for both conditions:

\begin{quote}\ttfamily\small
\#\# Consultation Context\\
\{context\}
\end{quote}

where \texttt{\{context\}} is the interleaved patient—clinician transcript up to and including the midpoint clinician turn, formatted as:

\begin{quote}\ttfamily\small
Clinician: \{utterance\}\\
Patient: \{utterance\}\\
...
\end{quote}

\subsubsection*{A.2.4 Canonical Response Selection Prompt}

Three candidate responses are generated per condition. A separate model call selects the most representative candidate for downstream evaluation:

\begin{quote}\ttfamily\small
You are choosing the most representative candidate response for downstream evaluation.

Select the single candidate that best represents the set as a typical output for the task. Return exact JSON only:
\{"choice": 1 | 2 | 3, "reasoning": "<1 sentence>"\}
\end{quote}

\subsubsection*{A.2.5 Pairwise Preference judgment Prompt}

\begin{quote}\ttfamily\small
You are an expert evaluator of clinical consultation quality.

\#\# Consultation Context (first half of consultation)\\
\{context\}

\#\# Ground Truth Next Three Clinician Turns\\
\{ground\_truth\}

\#\# Response A\\
\{response\_a\}

\#\# Response B\\
\{response\_b\}

Which response better reflects the patient's current emotional state and is more clinically appropriate as the next turn in this consultation?

Respond in this exact JSON format:\\
\{\\
\quad "preference": "A" | "B" | "TIE",\\
\quad "confidence": 1 | 2 | 3,\\
\quad "reasoning": "<2-3 sentences>"\\
\}\\
Use the full scale of judgments. Do not treat ties or extremely strong preferences as defaults. Do not include any text outside the JSON.
\end{quote}

Condition-to-label assignment (A/B) is randomized per consultation using a deterministic seed combining a global seed and the consultation identifier, ensuring reproducibility while balancing assignment across the dataset. The ground truth is presented as a reference anchor representing the actual clinician's next three turns; judges are not instructed to prefer responses closer to ground truth, but rather to use it as contextual scaffolding for assessing clinical appropriateness.

\subsubsection*{A.2.6 Absolute Rubric Scoring Prompt}

Each response is scored independently on four dimensions via separate model calls (one call per dimension per response):

\begin{quote}\ttfamily\small
You are an expert evaluator of clinical consultation quality.

\#\# Consultation Context\\
\{context\}

\#\# Ground Truth Next Three Clinician Turns\\
\{ground\_truth\}

\#\# Response to Evaluate\\
\{response\}

Rate the response on the following dimension only:

Dimension: \{dimension\_name\}\\
Description: \{dimension\_description\}

Scoring guidance:\\
- Use the full 1--5 scale.\\
- A score of 5 should be reserved for responses that satisfy the dimension exceptionally well with little or no meaningful room for improvement.\\
- Strong but imperfect responses should usually receive a 4 rather than a 5.\\
- Do not avoid giving a 5 when it is clearly warranted.

Respond in this exact JSON format:\\
\{"score": <integer 1--5>, "reasoning": "<1-2 sentences>"\}\\
Do not include any text outside the JSON.
\end{quote}

The four dimensions and their descriptions are given in Table~\ref{tab:rubric_dimensions}.

\begin{table}[h]
\centering
\caption{Rubric dimensions used in absolute scoring.}
\label{tab:rubric_dimensions}
\resizebox{\columnwidth}{!}{
\begin{tabular}{lp{9cm}}
\toprule
\textbf{Dimension} & \textbf{Description} \\
\midrule
Affective Attunement &
    Does the response accurately acknowledge the patient's current emotional state? \\
Clinical Appropriateness &
    Is this a response a competent clinician would give at this point in the consultation? \\
Contextual Coherence &
    Does the response follow naturally from the conversation history? \\
Specificity &
    Is the response specific to this patient, or generic and templated? \\
\bottomrule
\end{tabular}}
\end{table}

\subsubsection*{A.2.7 Ground Truth Alignment Scoring Prompt}

\begin{quote}\ttfamily\small
Rate how closely Response [\{label\}] matches the Ground Truth in terms of communicative intent and emotional attunement, on a scale of 1--5. You are not evaluating whether the ground truth is optimal—only measuring alignment with it. Treat the ground truth as a short reference trajectory across the next three clinician turns, not as a word-for-word target.

\#\# Ground Truth\\
\{ground\_truth\}

\#\# Response\\
\{response\}

Scoring guidance:\\
- Use the full 1--5 scale.\\
- Reserve a 5 for near-matches in communicative intent and emotional attunement.\\
- Good but not especially close matches should usually receive a 4 rather than a 5.\\
- Do not avoid giving a 5 when the match is truly very close.

Respond in this exact JSON format:\\
\{"score": <integer 1--5>, "reasoning": "<1 sentence>"\}
\end{quote}

\end{document}